\pdfoutput=1

\documentclass[11pt]{article}

\usepackage[]{acl}

\usepackage{times}
\usepackage{latexsym}
\usepackage{graphicx} 
\usepackage{float}
\usepackage{CJK}
\usepackage{soul}
\PassOptionsToPackage{prologue,dvipsnames}{xcolor}
\usepackage[dvipsnames]{xcolor}

\definecolor{c1}{HTML}{778d45}
\definecolor{c2}{HTML}{b71a3b}

\usepackage[T1]{fontenc}

\usepackage[utf8]{inputenc}

\usepackage{microtype}

\usepackage{inconsolata}
\usepackage{booktabs}
\usepackage{multirow}

\title{Towards Detecting LLMs Hallucination via \\ Markov Chain-based Multi-agent Debate Framework}

\author{Xiaoxi Sun$^{1}$\thanks{\ \ Equal contribution.}, Jinpeng Li$^{1}$\footnotemark[1], Yan Zhong$^{1}$, Dongyan Zhao$^{1}$\thanks{\ \ Corresponding authors: Dongyan Zhao and Rui Yan.}, Rui Yan$^{2}$\footnotemark[2]\\
$^1$Wangxuan Institute of Computer Technology, Peking University\\
$^2$Gaoling School of Artifical Intelligence, Renmin University of China\\
\texttt{\{sunxiaoxi, zhongyan\}@stu.pku.edu.cn, lijp.pku@gmail.com,} \\
\texttt{zhaody@pku.edu.cn, ruiyan@ruc.edu.cn}\\
}

\begin{document}
\maketitle
\begin{abstract}
The advent of large language models (LLMs) has facilitated the development of natural language text generation. 
It also poses unprecedented challenges, with content hallucination emerging as a significant concern. 
Existing solutions often involve expensive and complex interventions during the training process. Moreover, some approaches emphasize problem disassembly while neglecting the crucial validation process, leading to performance degradation or limited applications. 
To overcome these limitations, we propose a Markov Chain-based multi-agent debate verification framework to enhance hallucination detection accuracy in concise claims. Our method integrates the fact-checking process, including claim detection, evidence retrieval, and multi-agent verification. In the verification stage, we deploy multiple agents through flexible Markov Chain-based debates to validate individual claims, ensuring meticulous verification outcomes. 
Experimental results across three generative tasks demonstrate that our approach achieves significant improvements over baselines.

\end{abstract}

\section{Introduction}
The continuous evolution of large language models (LLMs) has significantly expanded language processing capabilities across diverse domains~\cite{wei2022emergent}. However, this progress introduces challenges, such as the substantial cost associated with updating model parameters and inherent deficiencies in reasoning~\cite{10.1145/3571730, zhang2023sirens, zheng2023does}. This has led to the generation of inaccurate content, known as hallucination, particularly concerning potent yet opaque models like ChatGPT and GPT-4~\cite{openai2023gpt4}.

\begin{figure}[t]
    \centering
    \includegraphics[scale=0.5]{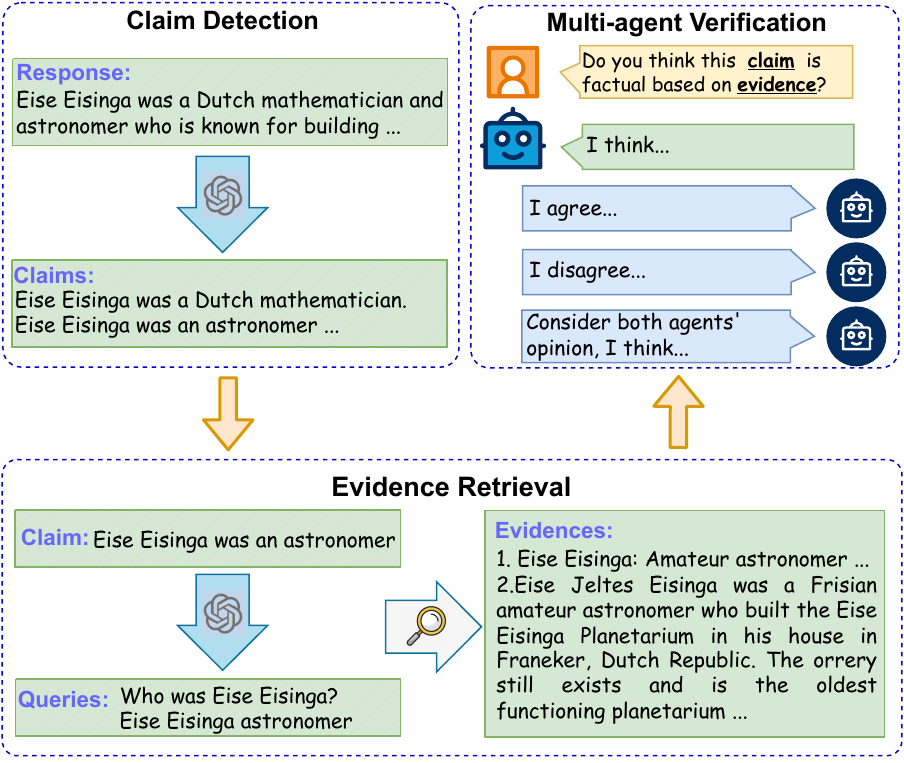}
    \caption{Overview of the fact-checking process, which includes three distinct stages. \textbf{Claim Detection}, we leverage large language models such as ChatGPT to obtain different claims. \textbf{Evidence Retrieval}, we prompt ChatGPT to formulate two queries, subsequently leveraging these queries to retrieve evidence with Google API or the provided knowledge. \textbf{Multi-agent Verification}, we propose the Markov Chain-based multi-agent debate verification framework, which can emulate human behavior to enhance model outputs and improve reasoning capabilities. Our main contributions focus on the process of Multi-agent Verification.
    }
    \label{Fig1}
\end{figure}

Hallucination detection has become a focal point in addressing these challenges. 
Existing methods often necessitate costly and intricate interventions during the training process~\cite{NEURIPS2022_df438caa, touvron2023llama, elaraby2023halo, wu2023finegrained}, rendering them unsuitable for large language models with agnostic parameters and these methods often incur considerable costs. 
Consequently, researchers have explored post-processing approaches~\cite{gao-etal-2023-rarr, peng2023check, chern2023factool, vu2023freshllms, gero2023selfverification} involving hallucination detection or correction post-content generation. 
Notably, these methods typically focus on problem decomposition and evidence retrieval, emphasizing simple prompting during individual verification. We posit that the verification accuracy is pivotal compared to problem decomposition in LLMs.

To address these challenges, we present a fact-checking process to enhance the accuracy of hallucination detection.
As shown in Figure~\ref{Fig1}, which involves three stages: claim detection, evidence retrieval, and multi-agent verification. 
In claim detection, our approach involves the extraction of claims from extensive responses by prompting ChatGPT, decomposing the intricate problem into smaller components.
Evidence retrieval involves generating queries based on claims for retrieval. Subsequently, we retrieve the corresponding evidence based on these generated queries. In the multi-agent verification stage, we innovatively propose a Markov Chain-based multi-agent debate verification framework, which leverages the robust capabilities of multi-agent systems to simulate human behavior.
This approach involves deploying diverse agents in Markov Chain debates to verify individual claims, thus providing a nuanced and flexible validation process. Following the verification of each claim using our method, the collective judgment of all claims contributes to the detection of hallucinations in the original response.

We conduct extensive experiments across three generative tasks, including question-answering, Summarization, and Dialogue, demonstrating the effectiveness of our approach. Verification outcomes are meticulously analyzed and compared against existing methods to ascertain the superiority of our approach. In summary, our contributions can be summarized as follows:

\begin{itemize}
\item We propose a versatile hallucination detection process applicable to multiple generation tasks for improving verification accuracy. 
\item We introduce a Markov Chain-based multi-agent debate verification framework that simulates human discussion.
\item Experiments conducted on three generative tasks show that our proposed framework outperforms baselines.
\end{itemize}

\begin{figure*}[t]
    \centering
    \includegraphics[scale=0.65]{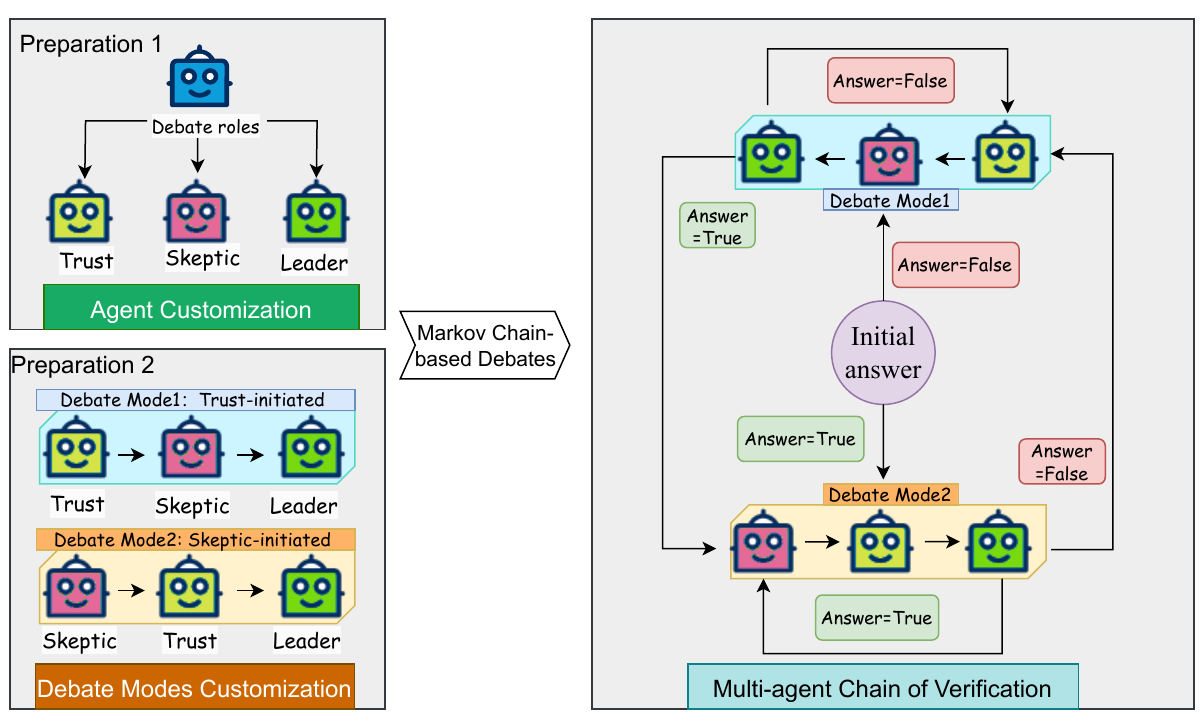}
    \caption{Overview of the proposed multi-agent debate verification framework for hallucination detection. There are two preparations before multi-agent debate verification. In preparation 1 (\textbf{Agent Customization}), we define three different debate agent roles, including Trust, Skeptic, and Leader roles. In preparation 2 (\textbf{Debate Modes Customization}), we assume that the debate process consists of two modes: Trust agent-initiated discussion (Trust-Skeptic-Leader) and Skeptic agent-initiated discussion (Skeptic-Trust-Leader). Then, in \textbf{Multi-agent Chain of Verification}, our validation process can be regarded as a Markov chain, which continually oscillates between these two debate modes to arrive at an optimal judgment.
    }
    \label{Fig2}
\end{figure*}

\section{Related Work}

\subsection{Hallucination Detection}





Before the emergence of large language models, hallucination detection was a significant topic within the field of natural language processing. Previous efforts primarily focused on detecting hallucinations in various tasks such as summarization \cite{kryscinski-etal-2020-evaluating,maynez-etal-2020-faithfulness, goyal-durrett-2021-annotating}, dialogue \cite{das-etal-2022-diving}, question-answering \cite{longpre-etal-2021-entity}, and machine translation \cite{xu-etal-2023-understanding}. These approaches primarily aimed to identify discrepancies between the generated content and the input, as well as internal inconsistencies within the generated content. However, they were often tailored specifically to task-specific models, lacking generalizability. 
There were also fact-checking endeavors that aimed to identify discrepancies between the generated content and real-world facts. This was typically accomplished through three steps \cite{guo-etal-2022-survey}: Claim Detection, Evidence Retrieval, and Verdict Prediction. With the advent of large language models, some works \cite{gao-etal-2023-rarr, li2023halueval} tackled the task of hallucination detection by prompting the large language models directly. In addition to task-specific approaches, there are hallucination detection methods specifically designed for LLMs. For example, some methods assess hallucination detection by examining the consistency of sampled examples \cite{manakul2023selfcheckgpt, zhang2023sac3}. Our work is fundamentally based on the fact-check framework. We transfer the Verdict Prediction stage to the Multi-agent Verification to improve the precision of validation.

\subsection{Hallucination Mitigation}

LLMs have demonstrated significant potential recently. However, they have not been able to completely eliminate the occurrence of hallucinations \cite{zheng2023does}. The extended text produced by these large models, which encompasses more diverse content and often introduces external knowledge, renders traditional methods for hallucination mitigation less effective. Consequently, a plethora of works dedicated to addressing hallucination mitigation in LLMs have emerged. 
Various approaches are presented to mitigate hallucinations at different stages of their application within the LLM life cycle \cite{zhang2023sirens}, including the pre-training phase of large models \cite{NEURIPS2022_df438caa, touvron2023llama}, the SFT phase \cite{chen2023alpagasus, elaraby2023halo}, the alignment phase \cite{wu2023finegrained, casper2023open}, and the decoding phase \cite{li2023inferencetime, chuang2023dola}. Implementing these methods necessitates adjustments to the model's parameters and requires a certain amount of training data, incurring some overhead. Numerous endeavors have been undertaken to mitigate hallucinations in content generated by black-box models, such as leveraging external knowledge bases or tools \cite{gao-etal-2023-rarr, peng2023check, chern2023factool, vu2023freshllms} and adopting self-refining approaches \cite{gero2023selfverification}. Our approach also centers on hallucination mitigation for black-box models, introducing a distinctive multi-agent method to augment its effectiveness.
\subsection{Multi-agent in LLMs}
In recent years, there has been a significant increase in the size of models and the amount of training data used, resulting in the exceptional performance of large language models (LLMs) across various tasks. As a result, researchers have explored the use of LLMs as agents to simulate human behavior, leading to the development of influential projects such as Generative Agents \cite{park2023generative}, Ghost in the Minecraft \cite{zhu2023ghost}, GPT-Bargaining \cite{fu2023improving} and Werewolf game \cite{xu2023exploring}. There are also some efforts involve multiple agents engaging in debates to improve the reasoning capabilities \cite{liang2023encouraging, du2023improving, xiong2023examining} or address issues related to hallucinations \cite{du2023improving,cohen2023lm}. However, existing methods for hallucination detection and mitigation of LLMs solely rely on natural language interactions between agents, which may pose concerns regarding the self-correction approach \cite{huang2023large}. Therefore, the objective of our work is to facilitate flexible discussions among multiple agents based on existing facts, aiming to detect and mitigate hallucinations in the generated content of language models.

\section{Method}
The primary objective of our study is to detect hallucinations in the content generated by the model. To accomplish this, we adhere to the conventional fact-checking process and make some modifications. The process is structured into three distinct stages: \textit{Claim Detection, Evidence Retrieval, and Multi-agent Verification}. This systematic approach enables the dissection of a complex problem into more manageable components. It has come to our attention that in certain fact-checking procedures, despite the accurate extraction of claims and the acquisition of robust evidence, verification errors persist in the final stage, undermining the efficacy of preceding efforts. 

Therefore, we propose a novel multi-agent debate verification framework for hallucination detection, the overview of which is shown in Figure~\ref{Fig2}. An anthropomorphic debate process based on the Markov chain is designed to be applicable across various generative tasks in the proposed method, bolstering the accuracy of verification. Subsequent sections will expound on each of these three stages individually, with a particular emphasis on our innovative approach in the third stage.
\subsection{Claim Detection}
In the stage of claim detection, we employ the methodology utilized in Factool \cite{chern2023factool}, leveraging large language models such as ChatGPT. Harnessing the robust instruction-following capabilities of LLMs empowers us to address the challenge of dissecting intricate responses. Nevertheless, detecting the hallucinations in statements lacking adequate information is futile and could impede overall judgment. Moreover, specific tasks may demand the concatenation of the model's responses with particular input information to formulate an informative claim, necessitating supplementary processing. Detailed explanations of these processing methods are provided in the experimental implementation section \S \ref{sec:implementation}.
\subsection{Evidence Retrieval}
Upon extraction of claims, a retrieval methodology is employed to ascertain corresponding evidence. Drawing inspiration from Factool's \cite{chern2023factool}  strategy in Knowledge Base Question Answering (KBQA) tasks, we prompt ChatGPT to formulate two queries, subsequently leveraging these queries to retrieve evidence. In instances where pertinent knowledge is absent, we employ the Google API to retrieve data from the internet. Conversely, when dealing with data accompanied by provided knowledge, we either consider the length of the knowledge as direct evidence or encode it for local retrieval.
\subsection{Multi-agent Verification}
We propose a Markov Chain-based multi-agent debate verification framework. Our investigations reveal significant potential in employing multi-agent systems to emulate human behavior \cite{park2023generative, zhu2023ghost}, particularly in the domain of fact-checking claims grounded in evidence. The effectiveness of addressing this task is notably heightened through the use of multi-agent debates. Despite considerable advancements in leveraging multi-agent debates to enhance model outputs and improve reasoning capabilities \cite{liang2023encouraging, du2023improving}, two critical aspects remain underexplored within the realm of hallucination detection.
\begin{itemize}
    \item[1)]
\textbf{Application to Verification:} Few studies have directly applied the multi-agent approach to the task of verification, they more concentrate on the decomposition of the complicated samples. Recognizing this research gap, our work aims to bridge it by introducing the multi-agent debate verification framework.
    \item[2)]
\textbf{Flexible Debate Process:} Existing methods in debate often adhere to a fixed process, unlike human debates where participants dynamically adjust their arguments based on prior outcomes. Our proposed approach takes inspiration from the Markov chain, where the selection of the current state depends on the results of a limited set of preceding states. This debate mode is more similar to the discussion between humans.
\end{itemize}

In summary, our multi-agent debate verification framework ingeniously adapts the multi-agent paradigm to the hallucination detection task. By infusing flexibility into the debate process and drawing inspiration from the Markov chain, our goal is to enhance the accuracy and adaptability of the verification process when assessing the veracity of claims based on evidence.

The key point in our method lies in the definition of states and the transition mechanisms.
\subsubsection{States}
\paragraph{Agents}
To comprehend the definition of states, it is imperative to elucidate the roles assumed by the diverse agents under consideration. We engage three distinct agents: \textit{Trust, Skeptic,} and \textit{Leader}. These agents collectively share the commonality of assimilating perspectives from one or more antecedent agents. They meticulously scrutinize these perspectives, grounded in claims and evidence amassed in preceding sections, express concurrence or dissent, and proffer their own viewpoints accompanied by factual assessments of the claims. The differentiation among these agents lies in their inclination toward antecedent viewpoints. The \textbf{Trust} agent predominantly leans towards accepting the perspectives of the preceding agent, thereby bolstering their credibility. Conversely, the \textbf{Skeptic} agent challenges the perspectives of the antecedent agent, diligently seeking to pinpoint inconsistencies between viewpoints and supporting evidence. The \textbf{Leader} agent amalgamates the perspectives of two agents, critically examines the rational and irrational facets, and ultimately formulates its own viewpoint. We implement agents with different personas through various prompts. Details can be found in Appendix \ref{sec:prompts}. The configuration of these agents, arranged in various sequences, constitutes the states delineated in our approach.
\paragraph{States}
We need to precisely define the states mentioned earlier. According to the definition of the Markov chain, we require an initial state to initiate our verification chain. Each agent must analyze the perspectives of preceding agents, necessitating an initial agent to furnish the primary answer for subsequent debate. This initial state is characterized by the initial agent, labeled as $S_{0}$, and our verification chain unfolds from this state.

We predominantly have two ordinary states, each comprising three agents. These states can be regarded as two distinct discussion modes. The first is the Trust agent-initiated discussion, labeled as $S_{1}$, following the sequence \textit{\textbf{Trust-Skeptic-Leader}}. This mode aims to bolster the credibility of the preceding viewpoint before introducing skepticism. The second state is initiated by the Skeptic agent, designated as $S_{2}$, with the sequence \textit{\textbf{Skeptic-Trust-Leader}}. This mode leans towards questioning the credibility of the previous viewpoints before further analyzing the skeptical perspective. Our verification chain continually oscillates between these two debate modes to arrive at an optimal judgment.

To prevent the chain from infinitely extending, a termination state is essential. Analogous to human debates concluding when opinions align, our termination condition is similar. If, within a state, the three agents reach a consensus, the chain terminates. When the Skeptic agent fails to identify points of contention, and the Leader, after scrutinizing their opinions, has no objections, yielding the same judgment, we consider the debate concluded. Additionally, we have imposed a maximum limit on verification rounds to constrain the length of the chain.
\subsubsection{Transition}
Transitioning between states is a critical aspect of our methodology, following the definition of states. The primary criterion guiding these transitions in our approach is the verification result of a claim by the preceding state. This methodology aligns with human intuition, acknowledging the potential for diverse perspectives in debating a given matter.

Our transition probabilities are as follows:
\begin{equation}
  Pr\left ( S_2 | R = True \right )  = 1
\end{equation}
\begin{equation}
      Pr\left ( S_1 | R = False \right )  = 1 
\end{equation}

$R$ represents the judgment obtained from the previous state. Specifically, our chosen transition method operates as follows: if the preceding state deems the current claim as factual, we transition to $S_{2}$. Our objective is to engage in a rigorous discussion, analyzing and questioning the claim only in the absence of contradictions in the previous state. The goal is to identify and address potential loopholes. If none are found, the Trust agent can reasonably conclude acceptance of the answer, leading to the convergence of the entire chain. Conversely, when the preceding state categorizes the claim as non-factual, we transition to $S_{1}$. In essence, we initially reinforce the credibility of this judgment, confirming the validity of skepticism. By enhancing the credibility of this opinion, if subsequent skepticism from the Skeptic agent is challenging, we can reasonably conclude the accuracy of this judgment, leading to the convergence of the chain.

Therefore, our overall process unfolds in the following manner: Initially, an initial answer is obtained from the initial state $S_{0}$. Based on this answer, the first transition to either $S_{1}$ or $S_{2}$ is made. Subsequent transitions rely solely on the judgment of the preceding state, continuing until a consensus is reached among the three agents within a state, culminating in the final verification result.
\section{Experiments}
We conducted experiments encompassing three generative tasks: Knowledge-Based Question Answering (KB-QA), Dialogue, and Summarization.

\subsection{Experimental Setup}
For all three tasks, we prompt the ChatGPT to execute claim extraction, query generation, and multi-agent debate verification. The verification process is iterated a minimum of 2 rounds, and 10 snippets of evidence are extracted. The chosen transition method involved switching to the skeptic agent when the response was determined to be True.
\subsubsection{Datasets and Baselines}
\label{sec:datasetandbaseline}
In this paper, we perform experiments on three different tasks, including Question-Answer (QA), Summarization, and Dialogue. The experimental datasets are derived from the following two canonical databases: 
\begin{itemize}
    \item Factool \cite{chern2023factool}: The Factprompts data comprises real-world questions with responses generated by ChatGPT, along with Factool-annotated claims extracted from these responses. 
    \item HaluEval \cite{li2023halueval}: HaluEval constitutes a substantial collection of sampling-then-filtering generated and human-annotated hallucinated samples, serving as an evaluation metric for language model performance in recognizing hallucination.
\end{itemize}

We randomly selected 150, 50, and 150 samples from the three tasks of HaluEval for testing purposes. The selection of samples was contingent upon the complexity of task responses, with summarization outputs being more intricate. Owing to the necessity of decomposing summarization into a greater number of claims, the extracted quantity is comparatively smaller than that of the other two tasks. The positive and negative instances within the dataset were randomly sampled using a binary distribution with a probability of 0.5. The resulting data distribution is presented in Table \ref{tab1}.

We compared the Factool method, the few-shot prompting method in HaluEval, the self-check method \cite{chern2023factool}, and our approach.

\begin{table}[t] 
\small
\centering
\begin{tabular}{lcc} \hline 
\textbf{Datasets}  & \textbf{Positive} & \textbf{Negative}   \\ \hline 

    Factool QA & 23 & 27 \\ \hline 

    HaluEval QA & 75 & 75  \\ \hline 

    HaluEval Summarization & 25 & 25 \\ \hline 

    HaluEval Dialogue & 80 &  70  \\ \hline

\end{tabular}
\caption{The Number of positive and negative samples in different datasets.}
\label{tab1}
\end{table}

\begin{table*}[t] 
\small
\centering
\vspace{5mm}
\setlength\tabcolsep{2pt}
\begin{tabular}{l@{\hspace{1em}}c@{\hspace{1em}}c@{\hspace{1em}}c@{\hspace{1em}}c@{\hspace{1em}}c@{\hspace{1em}}c@{\hspace{1em}}c@{\hspace{1em}}c}
\toprule  
\multirow{2}{*}{\textbf{Method}} & \multicolumn{4}{c}{\textbf{Claim-Level}} & \multicolumn{4}{c}{\textbf{Response-Level}} \\
 \cmidrule(l){2-5} \cmidrule(l){6-9}& \textbf{Acc.} & \textbf{R} & \textbf{P} & \textbf{F1}  & \textbf{Acc.} & \textbf{R} & \textbf{P} & \textbf{F1}  \\
\midrule
    Self-Check (0) & 75.54 & \textbf{90.40} & 80.00 & \textbf{84.88 }& 54.00 & \textbf{60.87} &  50.00 & 54.90 \\
    Self-Check (3) & 69.53 & 81.36 & 79.12 & 80.23 & 54.00 & 47.83 & 50.00 & 48.89 \\
    FACTOOL & 74.25 & 73.45 & \textbf{90.91} & 81.25 & 64.00 & 43.48 & 66.67 & 52.63 \\
    Our Method & \textbf{77.68} & 80.79 & 88.82 & 84.62 & \textbf{72.00} & 52.17 & \textbf{80.00} & \textbf{63.15} \\
\bottomrule 
\end{tabular}
\caption{Accuracy(\%), Recall(\%), Precision(\%), F1(\%) of four methods on Dataset Factool \cite{chern2023factool} used. Claim-Level denotes the results evaluated on all annotated claims, and Response-Level denotes the results evaluated on the origin responses. The best scores are highlighted in \textbf{bold}. }
\label{tab2}
\end{table*}

\begin{table*}[t]
\small
\centering
\vspace{5mm}
\setlength\tabcolsep{5pt}
\begin{tabular}{lcccccccccccc}
\toprule
\multirow{2}{*}{\textbf{Method}} & \multicolumn{4}{c}{\textbf{QA}} & \multicolumn{4}{c}{\textbf{Summarization}} & \multicolumn{4}{c}{\textbf{Dialogue}} \\ 
 \cmidrule(l){2-5} \cmidrule(l){6-9} \cmidrule(l){10-13}& \textbf{Acc.} & \textbf{R} & \textbf{P} & \textbf{F1}  & \textbf{Acc.} & \textbf{R} & \textbf{P} & \textbf{F1} & \textbf{Acc.} & \textbf{R} & \textbf{P} & \textbf{F1} \\
\midrule
    HaluEval & 56.00 & 77.33 & 54.21 & 63.74 & 58.00 & \textbf{100.0} & 54.35 & \textbf{70.42} & 68.00 & \textbf{75.71} & 63.10 & 68.83  \\ 
    FACTOOL & 67.33  & \textbf{86.67}  & 62.50 & 72.63 & 64.00 & 48.00 & 70.59 & 57.14 & 74.67 & 70.00 & 74,24 & \textbf{72.06}\\
    Ours & \textbf{70.67} & 82.67 & \textbf{66.67} & \textbf{73.81} & \textbf{70.00} & 64.00 & \textbf{72.73} & 68.09 & \textbf{76.00} & 62.86 & \textbf{81.48} & 70.97 \\
\bottomrule 
\end{tabular}
\caption{The results for our method and baseline on HaluEval \cite{li2023halueval} Dataset. We conducted experiments on three tasks: QA, Summarization, and Dialogue. The best scores are highlighted in \textbf{bold}.}
\label{tab3}
\end{table*}

\subsubsection{Implementataion Details}
\label{sec:implementation}
\paragraph{KB-QA}
For intricate and information-rich QA data, such as that in Factool \cite{chern2023factool}, we decomposed answers into multiple atomic claims and conducted multi-agent debate verification on each claim. If one of the claims is hallucinated, the origin answer is judged to be non-factual. As Factool data lacked corresponding evidence, Google search was employed to retrieve evidence for verification. In the case of simpler QA data, as found in HaluEval \cite{li2023halueval}, where answers sometimes are the single entity, such as \textit{"What American quartery lifestyle magazine did Hearst Shkelev Media also publish? Departures."}, we concatenated answers and questions to form QA pairs. Subsequently, we directly applied the multi-agent debate verification to these QA pairs, utilizing the provided knowledge in the dataset as evidence.

\paragraph{Summarization}
The model-generated summary was treated as a response, decomposed into multiple claims, and each claim was verified individually. The corresponding document to the summary served as evidence. To mitigate excessively long input queries, each sentence of the document was encoded separately, along with the query. The top 10 most similar sentences were selected as evidence for the current claim.

\paragraph{Dialogue}
In the course of the dialogue task, we encountered challenges associated with the extraction of claims. Dialogue responses frequently incorporated substantial subjective viewpoints such as \textit{"The last time that they made it to Super Bowl was in 2005. Are you a basketball fanatic too?"}, rendering the fact-checking of the factual accuracy of such subjective statements less meaningful. To mitigate this challenge, we introduced a pre-processing step wherein we directed ChatGPT to eliminate subjective portions from its responses prior to claim extraction, so the previous sentence becomes: \textit{"The last time that they made it to the Super Bowl was in 2005."}. This approach allowed us to retain only the informative segments for subsequent verification. Additionally, in the verification process during claim extraction, we employed the dialogue history and external knowledge as supporting evidence.

\subsection{Performance Analysis}
\label{sec:analysis}

The experimental results are presented in Table \ref{tab2} and Table \ref{tab3}. Table \ref{tab2} shows the performance of our method on Factool \cite{chern2023factool}, presenting results at both the claim and response levels. According to Table \ref{tab2}, we can observe that our proposed method can consistently achieve optimal accuracy when compared to various approaches. 

Table \ref{tab3} displays the test results on the HaluEval \cite{li2023halueval} dataset, from which we can observe that: Our method demonstrates optimal accuracy, excelling in most metrics in all three tasks, Notably, in the three tasks of this dataset, our method exhibits a relatively low recall score. This can be attributed to our approach, which involves questioning claims verified as factual, thereby ensuring the precise detection of errors when claims are misclassified. However, this approach also results in misjudging some claims that inherently lack hallucinations as non-factual. This phenomenon is further elucidated in \S~\ref{section:transition}.

\subsection{Ablation Study}
\paragraph{Transition Methods}
\label{section:transition}

We assessed the impact of distinct transition methods. From the QA section of HaluEval \cite{li2023halueval}, we extract 80 samples to evaluate the impact of four transition methods: transitioning to $S_{2}$ when the preceding state deemed the current claim devoid of hallucination ($True \to \textbf{Skeptic}$), transitioning to $S_{1}$ when the preceding state deemed the current claim devoid of hallucination ($True \to \textbf{Trust}$), consistently transitioning to $S_{1}$ or $S_{2}$ irrespective of the preceding state's judgment about the claim. The results, presented in Table \ref{tab4}, reveal that $True \to \textbf{Skeptic}$ achieved optimal performance across three metrics. This is primarily attributed to the fact that this transition method endeavors to challenge claims deemed factual in the preceding state, subsequently scrutinizing for potential oversights. In accordance with the details presented in \S~\ref{sec:analysis}, this phenomenon results in a lower recall score than the $True \to \textbf{Trust}$ method, concurrently demonstrating an elevated precision value.

\begin{table}[t] 
\small
\centering
\vspace{2mm}
\setlength\tabcolsep{5pt}
\begin{tabular}{lcccc}
\toprule
\textbf{Method}  & \textbf{Acc.} & \textbf{R} & \textbf{P} & \textbf{F1}  \\
\midrule
    $Always \  \textbf{Skeptic}$ & 65.00 & 84.21 & 59.26 & 69.57 \\
    $Always \  \textbf{Trust}$ & 68.75 & 84.21 & 62.75 & 71.91 \\
    $True \to \textbf{Trust}$ & 67.50 & \textbf{89.47} & 60.71 & 72.34\\
    $True \to \textbf{Skeptic}$ & \textbf{70.00} & 86.84 & \textbf{63.46} & \textbf{73.33}   \\ 

\bottomrule 
\end{tabular}
\caption{\textbf{Comparison of Different Transition Methods.} We evaluate the influence of transition methods on 80 QA samples, setting the minimum debate rounds to 2. The best scores are highlighted in \textbf{bold}. }
\label{tab4}
\end{table}

\paragraph{Minimum Rounds of Debate}

\begin{figure}[t]
    \includegraphics[scale=0.48]{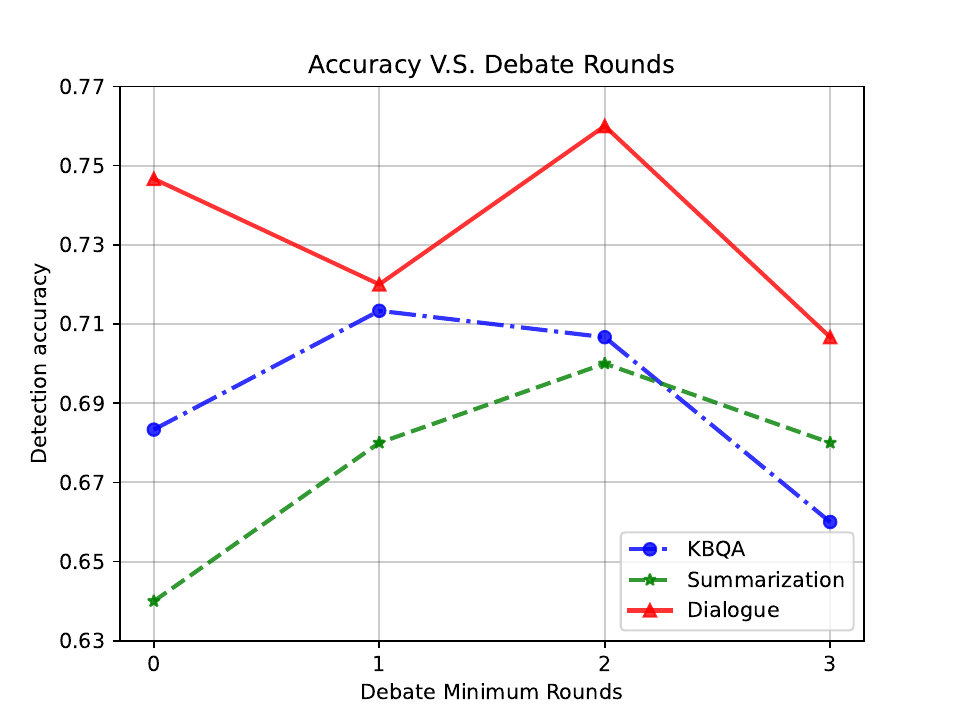}
    \caption{\textbf{Comparison of Different Minimum Debate Rounds.} We evaluate the influence of the minimum debate rounds on samples we utilized in \S \ref{sec:datasetandbaseline}, setting the transition method to $True \to \textbf{Skeptic}$. The \textbf{x}-axis represents different minimum debate rounds, whereas the \textbf{y}-axis signifies the corresponding detection accuracy.}
    \label{fig:debate-rounds}
\end{figure}

We explored the influence of different numbers of minimum debate rounds on the outcomes. We examined three distinct tasks using the previously extracted HaluEval data \cite{li2023halueval}, varying the number of minimum debate rounds from 0 to 3. Employing the $True \to \textbf{Skeptic}$ transition method, the results, illustrated in Figure \ref{fig:debate-rounds}, generally exhibit enhanced performance when the number of minimum rounds is set to 1 or 2, with a discernible decrease in efficacy when the number of minimum rounds is set to 3.

\paragraph{Comparison with Non-GPT Method}
In the multi-agent verification stage of the experiment in the Factool dataset, we employed the WeCheck\cite{wu-etal-2023-wecheck} method to conduct an ablation study, showcasing the benefits of our approach. We held the initial two steps constant, utilizing the Factool method to extract claims and retrieve evidence. Employing the claim as the hypothesis and the evidence as the premise, instances with WeCheck scores greater than or equal to 0.5 were deemed factual. From the experimental results in Table \ref{tab12}, we observed that compared to the non-GPT method, our approach exhibits significant advantages during the verification stage.

\begin{table}[t] 
\small
\centering
\vspace{2mm}
\setlength\tabcolsep{5pt}
\begin{tabular}{lcccc}
\toprule
\textbf{Method}  & \textbf{Acc.} & \textbf{R} & \textbf{P} & \textbf{F1}  \\
\midrule
    \textbf{Wecheck} & 65.23 & 64.41 & 86.36 & 73.78 \\
     \textbf{Our method} & \textbf{77.68} & \textbf{80.79} & \textbf{88.82} & \textbf{84.62} \\

\bottomrule 
\end{tabular}
\caption{\textbf{Comparison with Non-GPT Method.} We compare our method with Non-GPT Method Wecheck on the Factool dataset. The best scores are highlighted in \textbf{bold}. }
\label{tab12}
\end{table}

\subsection{Case Study}

To demonstrate the effectiveness of our approach, Table \ref{tab9} and Table \ref{tab10} show examples of the hallucination detection process for a Question-Answer (QA) sample. In Table \ref{tab9}, the debating agent is based on the GPT-3.5-turbo model, whereas Table \ref{tab10} utilizes GPT-4 as the base model.

When the debate starts, the initial agent generates an initial opinion based on the QA pair and the corresponding evidence. If no debate ensues, the initial opinion solidifies as the final answer. However, this approach overlooks both the insufficiency of evidence to support the claim that \textit{"The Landseer has a limited range of colors"} and the contradiction with evidence concerning \textit{"the English Mastiff having a wider range."} In Table \ref{tab9}, the three agents engage in discussions to highlight the insufficient evidence supporting \textit{"The Landseer has a limited range of colors."} However, they fail to infer the contradiction with the evidence suggesting \textit{"the English Mastiff has a wider range."} In Table \ref{tab10}, the agent, post-discussion, identifies both of these deficiencies. These observations indicate that larger language models, owing to their enhanced reasoning capabilities, yield better results when employing our method. Furthermore, it highlights that in some cases, a single round of debate may not reveal all inconsistencies between claim and evidence, emphasizing why sometimes increasing the minimum debate rounds can improve effectiveness.

\section{Conclusion}
In this paper, our purpose is to improve the accuracy of hallucination detection in content generated by large language models. Simultaneously, we aspired to extend this enhancement beyond particular generative tasks. To fulfill these objectives, we introduce a versatile framework for hallucination detection and propose the Markov Chain-based multi-agent debate verification framework. Our proposed approach demonstrates its effectiveness through evaluations conducted on both the Knowledge Base Question Answering (KBQA) dataset and the randomly sampled HaluEval dataset. We posit that our method demonstrates a level of generalizability, enabling its adaptation to other post-processing hallucination detection or mitigation approaches for better performance.

\section*{Limitations and Potential Risks}
Our methodology necessitates frequent interactions with the API of large language models (LLMs), resulting in significant overhead. This high frequency of API calls increases the cost and reduces response speed, which may limit its practicality in real-world scenarios. Nevertheless, this approach provides an accessible option for users lacking the infrastructure to implement large open-source models.

Furthermore, the distinctiveness among prompts for different agents primarily centers on role definition, while other aspects display considerable similarity. This occasionally leads to the partial repetition of opinions from the preceding agent. As exemplified by the two instances in Appendix \ref{sec:deb examples}, this phenomenon could be substantially alleviated by enhancing the performance of the base model.

\bibliography{anthology,custom}

\begin{thebibliography}{38}
\expandafter\ifx\csname natexlab\endcsname\relax\def\natexlab#1{#1}\fi

\bibitem[{Casper et~al.(2023)Casper, Davies, Shi, Gilbert, Scheurer, Rando, Freedman, Korbak, Lindner, Freire, Wang, Marks, Segerie, Carroll, Peng, Christoffersen, Damani, Slocum, Anwar, Siththaranjan, Nadeau, Michaud, Pfau, Krasheninnikov, Chen, Langosco, Hase, Bıyık, Dragan, Krueger, Sadigh, and Hadfield-Menell}]{casper2023open}
Stephen Casper, Xander Davies, Claudia Shi, Thomas~Krendl Gilbert, Jérémy Scheurer, Javier Rando, Rachel Freedman, Tomasz Korbak, David Lindner, Pedro Freire, Tony Wang, Samuel Marks, Charbel-Raphaël Segerie, Micah Carroll, Andi Peng, Phillip Christoffersen, Mehul Damani, Stewart Slocum, Usman Anwar, Anand Siththaranjan, Max Nadeau, Eric~J. Michaud, Jacob Pfau, Dmitrii Krasheninnikov, Xin Chen, Lauro Langosco, Peter Hase, Erdem Bıyık, Anca Dragan, David Krueger, Dorsa Sadigh, and Dylan Hadfield-Menell. 2023.
\newblock \href {http://arxiv.org/abs/2307.15217} {Open problems and fundamental limitations of reinforcement learning from human feedback}.

\bibitem[{Chen et~al.(2023)Chen, Li, Yan, Wang, Gunaratna, Yadav, Tang, Srinivasan, Zhou, Huang, and Jin}]{chen2023alpagasus}
Lichang Chen, Shiyang Li, Jun Yan, Hai Wang, Kalpa Gunaratna, Vikas Yadav, Zheng Tang, Vijay Srinivasan, Tianyi Zhou, Heng Huang, and Hongxia Jin. 2023.
\newblock \href {http://arxiv.org/abs/2307.08701} {Alpagasus: Training a better alpaca with fewer data}.

\bibitem[{Chern et~al.(2023)Chern, Chern, Chen, Yuan, Feng, Zhou, He, Neubig, and Liu}]{chern2023factool}
I-Chun Chern, Steffi Chern, Shiqi Chen, Weizhe Yuan, Kehua Feng, Chunting Zhou, Junxian He, Graham Neubig, and Pengfei Liu. 2023.
\newblock \href {http://arxiv.org/abs/2307.13528} {Factool: Factuality detection in generative ai -- a tool augmented framework for multi-task and multi-domain scenarios}.

\bibitem[{Chuang et~al.(2023)Chuang, Xie, Luo, Kim, Glass, and He}]{chuang2023dola}
Yung-Sung Chuang, Yujia Xie, Hongyin Luo, Yoon Kim, James Glass, and Pengcheng He. 2023.
\newblock \href {http://arxiv.org/abs/2309.03883} {Dola: Decoding by contrasting layers improves factuality in large language models}.

\bibitem[{Cohen et~al.(2023)Cohen, Hamri, Geva, and Globerson}]{cohen2023lm}
Roi Cohen, May Hamri, Mor Geva, and Amir Globerson. 2023.
\newblock \href {http://arxiv.org/abs/2305.13281} {Lm vs lm: Detecting factual errors via cross examination}.

\bibitem[{Das et~al.(2022)Das, Saha, and Srihari}]{das-etal-2022-diving}
Souvik Das, Sougata Saha, and Rohini Srihari. 2022.
\newblock \href {https://doi.org/10.18653/v1/2022.findings-emnlp.48} {Diving deep into modes of fact hallucinations in dialogue systems}.
\newblock In \emph{Findings of the Association for Computational Linguistics: EMNLP 2022}, pages 684--699, Abu Dhabi, United Arab Emirates. Association for Computational Linguistics.

\bibitem[{Du et~al.(2023)Du, Li, Torralba, Tenenbaum, and Mordatch}]{du2023improving}
Yilun Du, Shuang Li, Antonio Torralba, Joshua~B. Tenenbaum, and Igor Mordatch. 2023.
\newblock \href {http://arxiv.org/abs/2305.14325} {Improving factuality and reasoning in language models through multiagent debate}.

\bibitem[{Elaraby et~al.(2023)Elaraby, Lu, Dunn, Zhang, Wang, Liu, Tian, Wang, and Wang}]{elaraby2023halo}
Mohamed Elaraby, Mengyin Lu, Jacob Dunn, Xueying Zhang, Yu~Wang, Shizhu Liu, Pingchuan Tian, Yuping Wang, and Yuxuan Wang. 2023.
\newblock \href {http://arxiv.org/abs/2308.11764} {Halo: Estimation and reduction of hallucinations in open-source weak large language models}.

\bibitem[{Fu et~al.(2023)Fu, Peng, Khot, and Lapata}]{fu2023improving}
Yao Fu, Hao Peng, Tushar Khot, and Mirella Lapata. 2023.
\newblock \href {http://arxiv.org/abs/2305.10142} {Improving language model negotiation with self-play and in-context learning from ai feedback}.

\bibitem[{Gao et~al.(2023)Gao, Dai, Pasupat, Chen, Chaganty, Fan, Zhao, Lao, Lee, Juan, and Guu}]{gao-etal-2023-rarr}
Luyu Gao, Zhuyun Dai, Panupong Pasupat, Anthony Chen, Arun~Tejasvi Chaganty, Yicheng Fan, Vincent Zhao, Ni~Lao, Hongrae Lee, Da-Cheng Juan, and Kelvin Guu. 2023.
\newblock \href {https://doi.org/10.18653/v1/2023.acl-long.910} {{RARR}: Researching and revising what language models say, using language models}.
\newblock In \emph{Proceedings of the 61st Annual Meeting of the Association for Computational Linguistics (Volume 1: Long Papers)}, pages 16477--16508, Toronto, Canada. Association for Computational Linguistics.

\bibitem[{Gero et~al.(2023)Gero, Singh, Cheng, Naumann, Galley, Gao, and Poon}]{gero2023selfverification}
Zelalem Gero, Chandan Singh, Hao Cheng, Tristan Naumann, Michel Galley, Jianfeng Gao, and Hoifung Poon. 2023.
\newblock \href {https://openreview.net/forum?id=SBbJICrglS} {Self-verification improves few-shot clinical information extraction}.
\newblock In \emph{ICML 3rd Workshop on Interpretable Machine Learning in Healthcare (IMLH)}.

\bibitem[{Goyal and Durrett(2021)}]{goyal-durrett-2021-annotating}
Tanya Goyal and Greg Durrett. 2021.
\newblock \href {https://doi.org/10.18653/v1/2021.naacl-main.114} {Annotating and modeling fine-grained factuality in summarization}.
\newblock In \emph{Proceedings of the 2021 Conference of the North American Chapter of the Association for Computational Linguistics: Human Language Technologies}, pages 1449--1462, Online. Association for Computational Linguistics.

\bibitem[{Guo et~al.(2022)Guo, Schlichtkrull, and Vlachos}]{guo-etal-2022-survey}
Zhijiang Guo, Michael Schlichtkrull, and Andreas Vlachos. 2022.
\newblock \href {https://doi.org/10.1162/tacl_a_00454} {A survey on automated fact-checking}.
\newblock \emph{Transactions of the Association for Computational Linguistics}, 10:178--206.

\bibitem[{Huang et~al.(2023)Huang, Chen, Mishra, Zheng, Yu, Song, and Zhou}]{huang2023large}
Jie Huang, Xinyun Chen, Swaroop Mishra, Huaixiu~Steven Zheng, Adams~Wei Yu, Xinying Song, and Denny Zhou. 2023.
\newblock \href {http://arxiv.org/abs/2310.01798} {Large language models cannot self-correct reasoning yet}.

\bibitem[{Ji et~al.(2023)Ji, Lee, Frieske, Yu, Su, Xu, Ishii, Bang, Madotto, and Fung}]{10.1145/3571730}
Ziwei Ji, Nayeon Lee, Rita Frieske, Tiezheng Yu, Dan Su, Yan Xu, Etsuko Ishii, Ye~Jin Bang, Andrea Madotto, and Pascale Fung. 2023.
\newblock \href {https://doi.org/10.1145/3571730} {Survey of hallucination in natural language generation}.
\newblock \emph{ACM Comput. Surv.}, 55(12).

\bibitem[{Kryscinski et~al.(2020)Kryscinski, McCann, Xiong, and Socher}]{kryscinski-etal-2020-evaluating}
Wojciech Kryscinski, Bryan McCann, Caiming Xiong, and Richard Socher. 2020.
\newblock \href {https://doi.org/10.18653/v1/2020.emnlp-main.750} {Evaluating the factual consistency of abstractive text summarization}.
\newblock In \emph{Proceedings of the 2020 Conference on Empirical Methods in Natural Language Processing (EMNLP)}, pages 9332--9346, Online. Association for Computational Linguistics.

\bibitem[{Lee et~al.(2022)Lee, Ping, Xu, Patwary, Fung, Shoeybi, and Catanzaro}]{NEURIPS2022_df438caa}
Nayeon Lee, Wei Ping, Peng Xu, Mostofa Patwary, Pascale~N Fung, Mohammad Shoeybi, and Bryan Catanzaro. 2022.
\newblock \href {https://proceedings.neurips.cc/paper_files/paper/2022/file/df438caa36714f69277daa92d608dd63-Paper-Conference.pdf} {Factuality enhanced language models for open-ended text generation}.
\newblock In \emph{Advances in Neural Information Processing Systems}, volume~35, pages 34586--34599. Curran Associates, Inc.

\bibitem[{Li et~al.(2023{\natexlab{a}})Li, Cheng, Zhao, Nie, and Wen}]{li2023halueval}
Junyi Li, Xiaoxue Cheng, Wayne~Xin Zhao, Jian-Yun Nie, and Ji-Rong Wen. 2023{\natexlab{a}}.
\newblock \href {http://arxiv.org/abs/2305.11747} {Halueval: A large-scale hallucination evaluation benchmark for large language models}.

\bibitem[{Li et~al.(2023{\natexlab{b}})Li, Patel, Viégas, Pfister, and Wattenberg}]{li2023inferencetime}
Kenneth Li, Oam Patel, Fernanda Viégas, Hanspeter Pfister, and Martin Wattenberg. 2023{\natexlab{b}}.
\newblock \href {http://arxiv.org/abs/2306.03341} {Inference-time intervention: Eliciting truthful answers from a language model}.

\bibitem[{Liang et~al.(2023)Liang, He, Jiao, Wang, Wang, Wang, Yang, Tu, and Shi}]{liang2023encouraging}
Tian Liang, Zhiwei He, Wenxiang Jiao, Xing Wang, Yan Wang, Rui Wang, Yujiu Yang, Zhaopeng Tu, and Shuming Shi. 2023.
\newblock \href {http://arxiv.org/abs/2305.19118} {Encouraging divergent thinking in large language models through multi-agent debate}.

\bibitem[{Longpre et~al.(2021)Longpre, Perisetla, Chen, Ramesh, DuBois, and Singh}]{longpre-etal-2021-entity}
Shayne Longpre, Kartik Perisetla, Anthony Chen, Nikhil Ramesh, Chris DuBois, and Sameer Singh. 2021.
\newblock \href {https://doi.org/10.18653/v1/2021.emnlp-main.565} {Entity-based knowledge conflicts in question answering}.
\newblock In \emph{Proceedings of the 2021 Conference on Empirical Methods in Natural Language Processing}, pages 7052--7063, Online and Punta Cana, Dominican Republic. Association for Computational Linguistics.

\bibitem[{Manakul et~al.(2023)Manakul, Liusie, and Gales}]{manakul2023selfcheckgpt}
Potsawee Manakul, Adian Liusie, and Mark J.~F. Gales. 2023.
\newblock \href {http://arxiv.org/abs/2303.08896} {Selfcheckgpt: Zero-resource black-box hallucination detection for generative large language models}.

\bibitem[{Maynez et~al.(2020)Maynez, Narayan, Bohnet, and McDonald}]{maynez-etal-2020-faithfulness}
Joshua Maynez, Shashi Narayan, Bernd Bohnet, and Ryan McDonald. 2020.
\newblock \href {https://doi.org/10.18653/v1/2020.acl-main.173} {On faithfulness and factuality in abstractive summarization}.
\newblock In \emph{Proceedings of the 58th Annual Meeting of the Association for Computational Linguistics}, pages 1906--1919, Online. Association for Computational Linguistics.

\bibitem[{OpenAI(2023)}]{openai2023gpt4}
OpenAI. 2023.
\newblock \href {http://arxiv.org/abs/2303.08774} {Gpt-4 technical report}.

\bibitem[{Park et~al.(2023)Park, O'Brien, Cai, Morris, Liang, and Bernstein}]{park2023generative}
Joon~Sung Park, Joseph~C. O'Brien, Carrie~J. Cai, Meredith~Ringel Morris, Percy Liang, and Michael~S. Bernstein. 2023.
\newblock \href {http://arxiv.org/abs/2304.03442} {Generative agents: Interactive simulacra of human behavior}.

\bibitem[{Peng et~al.(2023)Peng, Galley, He, Cheng, Xie, Hu, Huang, Liden, Yu, Chen, and Gao}]{peng2023check}
Baolin Peng, Michel Galley, Pengcheng He, Hao Cheng, Yujia Xie, Yu~Hu, Qiuyuan Huang, Lars Liden, Zhou Yu, Weizhu Chen, and Jianfeng Gao. 2023.
\newblock \href {http://arxiv.org/abs/2302.12813} {Check your facts and try again: Improving large language models with external knowledge and automated feedback}.

\bibitem[{Touvron et~al.(2023)Touvron, Martin, Stone, Albert, Almahairi, Babaei, Bashlykov, Batra, Bhargava, Bhosale, Bikel, Blecher, Ferrer, Chen, Cucurull, Esiobu, Fernandes, Fu, Fu, Fuller, Gao, Goswami, Goyal, Hartshorn, Hosseini, Hou, Inan, Kardas, Kerkez, Khabsa, Kloumann, Korenev, Koura, Lachaux, Lavril, Lee, Liskovich, Lu, Mao, Martinet, Mihaylov, Mishra, Molybog, Nie, Poulton, Reizenstein, Rungta, Saladi, Schelten, Silva, Smith, Subramanian, Tan, Tang, Taylor, Williams, Kuan, Xu, Yan, Zarov, Zhang, Fan, Kambadur, Narang, Rodriguez, Stojnic, Edunov, and Scialom}]{touvron2023llama}
Hugo Touvron, Louis Martin, Kevin Stone, Peter Albert, Amjad Almahairi, Yasmine Babaei, Nikolay Bashlykov, Soumya Batra, Prajjwal Bhargava, Shruti Bhosale, Dan Bikel, Lukas Blecher, Cristian~Canton Ferrer, Moya Chen, Guillem Cucurull, David Esiobu, Jude Fernandes, Jeremy Fu, Wenyin Fu, Brian Fuller, Cynthia Gao, Vedanuj Goswami, Naman Goyal, Anthony Hartshorn, Saghar Hosseini, Rui Hou, Hakan Inan, Marcin Kardas, Viktor Kerkez, Madian Khabsa, Isabel Kloumann, Artem Korenev, Punit~Singh Koura, Marie-Anne Lachaux, Thibaut Lavril, Jenya Lee, Diana Liskovich, Yinghai Lu, Yuning Mao, Xavier Martinet, Todor Mihaylov, Pushkar Mishra, Igor Molybog, Yixin Nie, Andrew Poulton, Jeremy Reizenstein, Rashi Rungta, Kalyan Saladi, Alan Schelten, Ruan Silva, Eric~Michael Smith, Ranjan Subramanian, Xiaoqing~Ellen Tan, Binh Tang, Ross Taylor, Adina Williams, Jian~Xiang Kuan, Puxin Xu, Zheng Yan, Iliyan Zarov, Yuchen Zhang, Angela Fan, Melanie Kambadur, Sharan Narang, Aurelien Rodriguez, Robert Stojnic, Sergey Edunov, and Thomas
  Scialom. 2023.
\newblock \href {http://arxiv.org/abs/2307.09288} {Llama 2: Open foundation and fine-tuned chat models}.

\bibitem[{Vu et~al.(2023)Vu, Iyyer, Wang, Constant, Wei, Wei, Tar, Sung, Zhou, Le, and Luong}]{vu2023freshllms}
Tu~Vu, Mohit Iyyer, Xuezhi Wang, Noah Constant, Jerry Wei, Jason Wei, Chris Tar, Yun-Hsuan Sung, Denny Zhou, Quoc Le, and Thang Luong. 2023.
\newblock \href {http://arxiv.org/abs/2310.03214} {Freshllms: Refreshing large language models with search engine augmentation}.

\bibitem[{Wei et~al.(2022)Wei, Tay, Bommasani, Raffel, Zoph, Borgeaud, Yogatama, Bosma, Zhou, Metzler, Chi, Hashimoto, Vinyals, Liang, Dean, and Fedus}]{wei2022emergent}
Jason Wei, Yi~Tay, Rishi Bommasani, Colin Raffel, Barret Zoph, Sebastian Borgeaud, Dani Yogatama, Maarten Bosma, Denny Zhou, Donald Metzler, Ed~H. Chi, Tatsunori Hashimoto, Oriol Vinyals, Percy Liang, Jeff Dean, and William Fedus. 2022.
\newblock \href {https://openreview.net/forum?id=yzkSU5zdwD} {Emergent abilities of large language models}.
\newblock \emph{Transactions on Machine Learning Research}.
\newblock Survey Certification.

\bibitem[{Wu et~al.(2023{\natexlab{a}})Wu, Li, Xiao, Liu, Li, and Lyu}]{wu-etal-2023-wecheck}
Wenhao Wu, Wei Li, Xinyan Xiao, Jiachen Liu, Sujian Li, and Yajuan Lyu. 2023{\natexlab{a}}.
\newblock \href {https://doi.org/10.18653/v1/2023.acl-long.18} {{W}e{C}heck: Strong factual consistency checker via weakly supervised learning}.
\newblock In \emph{Proceedings of the 61st Annual Meeting of the Association for Computational Linguistics (Volume 1: Long Papers)}, pages 307--321, Toronto, Canada. Association for Computational Linguistics.

\bibitem[{Wu et~al.(2023{\natexlab{b}})Wu, Hu, Shi, Dziri, Suhr, Ammanabrolu, Smith, Ostendorf, and Hajishirzi}]{wu2023finegrained}
Zeqiu Wu, Yushi Hu, Weijia Shi, Nouha Dziri, Alane Suhr, Prithviraj Ammanabrolu, Noah~A. Smith, Mari Ostendorf, and Hannaneh Hajishirzi. 2023{\natexlab{b}}.
\newblock \href {http://arxiv.org/abs/2306.01693} {Fine-grained human feedback gives better rewards for language model training}.

\bibitem[{Xiong et~al.(2023)Xiong, Ding, Cao, Liu, and Qin}]{xiong2023examining}
Kai Xiong, Xiao Ding, Yixin Cao, Ting Liu, and Bing Qin. 2023.
\newblock \href {http://arxiv.org/abs/2305.11595} {Examining inter-consistency of large language models collaboration: An in-depth analysis via debate}.

\bibitem[{Xu et~al.(2023{\natexlab{a}})Xu, Agrawal, Briakou, Martindale, and Carpuat}]{xu-etal-2023-understanding}
Weijia Xu, Sweta Agrawal, Eleftheria Briakou, Marianna~J. Martindale, and Marine Carpuat. 2023{\natexlab{a}}.
\newblock \href {https://doi.org/10.1162/tacl_a_00563} {Understanding and detecting hallucinations in neural machine translation via model introspection}.
\newblock \emph{Transactions of the Association for Computational Linguistics}, 11:546--564.

\bibitem[{Xu et~al.(2023{\natexlab{b}})Xu, Wang, Li, Luo, Wang, Liu, and Liu}]{xu2023exploring}
Yuzhuang Xu, Shuo Wang, Peng Li, Fuwen Luo, Xiaolong Wang, Weidong Liu, and Yang Liu. 2023{\natexlab{b}}.
\newblock \href {http://arxiv.org/abs/2309.04658} {Exploring large language models for communication games: An empirical study on werewolf}.

\bibitem[{Zhang et~al.(2023{\natexlab{a}})Zhang, Li, Das, Malin, and Kumar}]{zhang2023sac3}
Jiaxin Zhang, Zhuohang Li, Kamalika Das, Bradley~A. Malin, and Sricharan Kumar. 2023{\natexlab{a}}.
\newblock \href {http://arxiv.org/abs/2311.01740} {Sac$^3$: Reliable hallucination detection in black-box language models via semantic-aware cross-check consistency}.

\bibitem[{Zhang et~al.(2023{\natexlab{b}})Zhang, Li, Cui, Cai, Liu, Fu, Huang, Zhao, Zhang, Chen, Wang, Luu, Bi, Shi, and Shi}]{zhang2023sirens}
Yue Zhang, Yafu Li, Leyang Cui, Deng Cai, Lemao Liu, Tingchen Fu, Xinting Huang, Enbo Zhao, Yu~Zhang, Yulong Chen, Longyue Wang, Anh~Tuan Luu, Wei Bi, Freda Shi, and Shuming Shi. 2023{\natexlab{b}}.
\newblock \href {http://arxiv.org/abs/2309.01219} {Siren's song in the ai ocean: A survey on hallucination in large language models}.

\bibitem[{Zheng et~al.(2023)Zheng, Huang, and Chang}]{zheng2023does}
Shen Zheng, Jie Huang, and Kevin Chen-Chuan Chang. 2023.
\newblock \href {http://arxiv.org/abs/2304.10513} {Why does chatgpt fall short in providing truthful answers?}

\bibitem[{Zhu et~al.(2023)Zhu, Chen, Tian, Tao, Su, Yang, Huang, Li, Lu, Wang, Qiao, Zhang, and Dai}]{zhu2023ghost}
Xizhou Zhu, Yuntao Chen, Hao Tian, Chenxin Tao, Weijie Su, Chenyu Yang, Gao Huang, Bin Li, Lewei Lu, Xiaogang Wang, Yu~Qiao, Zhaoxiang Zhang, and Jifeng Dai. 2023.
\newblock \href {http://arxiv.org/abs/2305.17144} {Ghost in the minecraft: Generally capable agents for open-world environments via large language models with text-based knowledge and memory}.

\end{thebibliography}

\newpage
\appendix
\section{Appendix}
\subsection{Prompts}
\label{sec:prompts}

Table \ref{tab5}, \ref{tab6}, \ref{tab7} and \ref{tab8} enumerate various prompts employed in our experimental design, including prompts for establishing different roles for the agent and prompts for eliminating subjective opinions from dialogue responses.
\begin{table*}[h!]
\begin{tabular}{p{\textwidth}}
\toprule
  You are the *Trust* agent of the three agents. Your task is to trust the previous agent's opinions as much as possible and to further expand upon it.

  You are given the opinions generated by a previous agent. Refer to the claim[text] and evidence[evidences] to analyse the opinions[previous opinions] of the previous agent. Carefully examine whether the corresponding evidences support the statements put forward by the previous agent. If you believe any part of them is accurate, please further analyze it based on that.   

  Then assess the factuality of the initial claim [text] based on the provided information [evidences] and the previous agent's opinions. 

  DO NOT REPEAT the previous agent's opinion, you should develop your own perspective based on theirs. Take the previous agent's opinions as a reference rather than copying it directly.

  The response should be a dictionary with three keys - "opinion", "factuality", "Error severity" which correspond to the your opinion, whether the given text is factual or not (Boolean - True or False), the severity level of claim error(Integer - Range from 0 to 5).

  The definitions of different levels of error severity are as follows:
  \begin{itemize}
    \item[0.] 
   \textbf{No Error (Grade 0)}: This is when the claim is entirely factual and accurate, with no errors or inaccuracies.
    \item[1.] 
   \textbf{Minor Errors (Grade 1)}: These are small and insignificant errors that do not significantly alter the essence or validity of the claim. For example, minor spelling mistakes, incorrect dates or small numerical discrepancies.
    \item[2.] 
   \textbf{Moderate Errors (Grade 2)}: These are errors that have some impact on the validity of the claim but do not reverse its overall meaning. For example, wrong terminologies, improper use of statistics or minor deviations from the evidence.
    \item[3.]
   \textbf{Major Errors (Grade 3)}: These are errors that have a significant impact on the validity of the claim. These errors may lead to significant misunderstandings or misinterpretations. Examples could be gross exaggeration or underestimation, misuse of expert authority, or manipulation of context.
    \item[4.] 
   \textbf{Critical Errors (Grade 4)}: These are errors that completely negate or invalidate the claim. The evidence contradicts the claim so fundamentally that its veracity is entirely undermined. For example, attribution of a quote or event to the wrong person or an incorrect recap of a major event.
    \item[5.] 
   \textbf{False Claim (Grade 5)}: This is when claims are entirely made up or intentionally deceptive, without any grounding in evidence. They are blatant lies designed to mislead or deceive, fabricating events, people, or statements that do not exist or never occurred.
\end{itemize}
  YOU SHOULD ONLY RESPONSE IN FORMAT AS DESCRIBED BELOW. DO NOT RETURN ANYTHING ELSE. START YOUR RESPONSE WITH '\{\{'.

  [response format]: 

  \{\{

   "opinion": "First analyze the opinions of the previous agent, point out what you think is correct or incorrect in its opinions and explain the reasons. Remember that you should trust previous agent's opinion as much as possible.  And then describe your opinion about factuality of the claim[text] based on the evidences [evidences]. Your opinions should be supported by corresponding evidences. DO NOT REITERATE THE PREVIOUS AGENT'S OPINIONS [previous opinions].",

   "factuality": True if the given text is factual, False otherwise.,

   "Error severity": Integer - Range from 0 to 5. The severity level of claim error. According to the definitions of levels of the error severity, please carefully compare "claim" and "evidence" and provide appropriate degree.

  \}\}
\\ \bottomrule
\end{tabular}
\caption{Prompt For Trust Agent}
\label{tab5}
\end{table*}

\begin{table*}[h!]
\begin{tabular}{p{\textwidth}}
\toprule
You are the *Skeptic* agent of the three agents. "Skeptic" means you must question the previous agent's opinions by scrutinizing the available data [text] and [evidences] and identifying any possible errors or misleading factors in previous agent's opinions.

You are given the opinions generated by a previous agent. Refer to the claim[text] and evidence[evidences] to analyse the opinions[previous opinions] of the previous agent. Carefully examine whether the corresponding evidences support the statements put forward by the previous agent. If you believe any part of its opinions is incorrect, please point it out and explain your perspective. 

Then critically examine the validity of the claim [text], considering any potential biases or inconsistencies between the information [evidences] and claim [text]. 

DO NOT REPEAT the previous agent's opinion, you should develop your own perspective based on theirs. Take the previous agent's opinions as a reference rather than copying it directly.

The response should be a dictionary with three keys - "opinion", "factuality", "Error severity" which correspond to the your opinion, whether the given text is factual or not (Boolean - True or False), the severity level of claim error(Integer - Range from 0 to 5).

  The definitions of different levels of error severity are as follows:
  \begin{itemize}
    \item[0.] 
   \textbf{No Error (Grade 0)}: This is when the claim is entirely factual and accurate, with no errors or inaccuracies.
    \item[1.] 
   \textbf{Minor Errors (Grade 1)}: These are small and insignificant errors that do not significantly alter the essence or validity of the claim. For example, minor spelling mistakes, incorrect dates or small numerical discrepancies.
    \item[2.] 
   \textbf{Moderate Errors (Grade 2)}: These are errors that have some impact on the validity of the claim but do not reverse its overall meaning. For example, wrong terminologies, improper use of statistics or minor deviations from the evidence.
    \item[3.]
   \textbf{Major Errors (Grade 3)}: These are errors that have a significant impact on the validity of the claim. These errors may lead to significant misunderstandings or misinterpretations. Examples could be gross exaggeration or underestimation, misuse of expert authority, or manipulation of context.
    \item[4.] 
   \textbf{Critical Errors (Grade 4)}: These are errors that completely negate or invalidate the claim. The evidence contradicts the claim so fundamentally that its veracity is entirely undermined. For example, attribution of a quote or event to the wrong person or an incorrect recap of a major event.
    \item[5.] 
   \textbf{False Claim (Grade 5)}: This is when claims are entirely made up or intentionally deceptive, without any grounding in evidence. They are blatant lies designed to mislead or deceive, fabricating events, people, or statements that do not exist or never occurred.
\end{itemize}
  YOU SHOULD ONLY RESPONSE IN FORMAT AS DESCRIBED BELOW. DO NOT RETURN ANYTHING ELSE. START YOUR RESPONSE WITH '\{\{'.

  [response format]: 

  \{\{

   "opinion": "First analyze the opinions of the previous agent, point out what you think is correct or incorrect in its opinions and explain the reasons. Remember that you should doubt the opinions of the previous agent as much as possible.  And then describe your opinion about factuality of the claim[text] based on the evidences [evidences]. Your opinions should be supported by corresponding evidences. DO NOT REITERATE THE PREVIOUS AGENT'S OPINIONS [previous opinions].",

   "factuality": True if the given text is factual, False otherwise.,

   "Error severity": Integer - Range from 0 to 5. The severity level of claim error. According to the definitions of levels of the error severity, please carefully compare "claim" and "evidence" and provide appropriate degree.

  \}\}

\\ \bottomrule
\end{tabular}
\caption{Prompt For Skeptic Agent}
\label{tab6}
\end{table*}

\begin{table*}[h!]
\begin{tabular}{p{\textwidth}}
\toprule
You are the *Leader* agent of the three agents. The other two agents are 'Trust' and 'Skeptic' agents. The 'Trust' agent will trust the previous agent's opinion as soon as possible and the 'Skeptic' agent will doubt the previous agent's opinion as soon as possible. 

You are given the opinions generated by two previous agents. Incorporate the opinions[previous opinions] provided by the 'Trust' and 'Skeptic' agents to synthesize the most accurate and reliable conclusion regarding the claim([text])'s veracity based on the [evidences]. You need to consider the characteristics of these two agents when forming your own opinions. Assess the strengths and weaknesses of both sides, and leverage the provided information to generate a conclusive judgment. 

DO NOT REPEAT the previous agent's opinion, you should develop your own perspective based on theirs. Take the previous agent's opinions as a reference rather than copying it directly.

The response should be a dictionary with three keys - "opinion", "factuality", "Error severity" which correspond to the your opinion, whether the given text is factual or not (Boolean - True or False), the severity level of claim error(Integer - Range from 0 to 5).

  The definitions of different levels of error severity are as follows:
  \begin{itemize}
    \item[0.] 
   \textbf{No Error (Grade 0)}: This is when the claim is entirely factual and accurate, with no errors or inaccuracies.
    \item[1.] 
   \textbf{Minor Errors (Grade 1)}: These are small and insignificant errors that do not significantly alter the essence or validity of the claim. For example, minor spelling mistakes, incorrect dates or small numerical discrepancies.
    \item[2.] 
   \textbf{Moderate Errors (Grade 2)}: These are errors that have some impact on the validity of the claim but do not reverse its overall meaning. For example, wrong terminologies, improper use of statistics or minor deviations from the evidence.
    \item[3.]
   \textbf{Major Errors (Grade 3)}: These are errors that have a significant impact on the validity of the claim. These errors may lead to significant misunderstandings or misinterpretations. Examples could be gross exaggeration or underestimation, misuse of expert authority, or manipulation of context.
    \item[4.] 
   \textbf{Critical Errors (Grade 4)}: These are errors that completely negate or invalidate the claim. The evidence contradicts the claim so fundamentally that its veracity is entirely undermined. For example, attribution of a quote or event to the wrong person or an incorrect recap of a major event.
    \item[5.] 
   \textbf{False Claim (Grade 5)}: This is when claims are entirely made up or intentionally deceptive, without any grounding in evidence. They are blatant lies designed to mislead or deceive, fabricating events, people, or statements that do not exist or never occurred.
\end{itemize}
  YOU SHOULD ONLY RESPONSE IN FORMAT AS DESCRIBED BELOW. DO NOT RETURN ANYTHING ELSE. START YOUR RESPONSE WITH '\{\{'.

  [response format]: 

  \{\{

   "opinion": "First explain your thoughts on the opinions of the two previous agents. And then describe your opinions about the factuality of the claim[text] based on the evidences [evidences]. Your opinions should be supported by corresponding evidences. DO NOT REITERATE ANY OF THE PREVIOUS AGENTS' OPINIONS [previous opinions]. With reference to the opinions of the "Trust" agent and the "Skeptic" agent, derive a novel opinion that you believe to be correct.",

   "factuality": True if the given text is factual, False otherwise.,

   "Error severity": Integer - Range from 0 to 5. The severity level of claim error. According to the definitions of levels of the error severity, please carefully compare "claim" and "evidence" and provide appropriate degree.

  \}\}

\\ \bottomrule
\end{tabular}
\caption{Prompt For Leader Agent }
\label{tab7}
\end{table*}

\begin{table*}[h!]
\begin{tabular}{p{\textwidth}}
\toprule

  You are given a piece of text, please remove sentences that you believe are totally personal opinions and do not contain any factual information.

  Your output should be the sentence after modification of the original content. If you believe the entire sentence is a personal opinion, please output None.

  Here are two examples: 
  
  [text]: My pleasure, let me know if you need more recommendations.

[response]: None \\

  [text]: The Fault in Our Stars is a fantasy novel. Have you read it? 

  [response]: The Fault in Our Stars is a fantasy novel. \\
  
  Now complete the following

[text]: \{input\}

  [response]: 

\\ \bottomrule
\end{tabular}
\caption{Prompt for extracting effective information}
\label{tab8}
\end{table*}

\subsection{Debate examples}
\label{sec:deb examples}

In Table \ref{tab9} and Table \ref{tab10}, we present two instances of Multi-Agent Debate Verification on a HaluEval QA sample. The agents involved in these two instances employ distinct base models: \textbf{GPT-3.5-turbo} and \textbf{GPT-4}. In Table \ref{tab11}, we provide a comprehensive breakdown of our verification method's inference process for better understanding, detailing the inputs and outputs of each agent.

\begin{table*}[h!]
\begin{tabular}{p{\textwidth}}
\toprule
\textbf{Question}: Does the Landseer or English Mastiff have a limited range of colors?\\

\textbf{Answer}: The Landseer has a limited range of colours, while the English Mastiff has a wider range.\\

\textbf{Evidence}: It is not to be confused with a white and black Newfoundland, which is also often called a landseer.The English Mastiff is a breed of extremely large dog (often known simply as the Mastiff) perhaps descended from the ancient Alaunt and Pugnaces Britanniae, with a significant input from the Alpine Mastiff in the 19th century. Distinguishable by enormous size, massive head, and a limited range of colours, but always displaying a black mask, the Mastiff is noted for its gentle and loving nature.\\

\textbf{Initial Opinion}: Based on the evidence provided, it is stated that the Landseer has a limited range of colors, while the English Mastiff has a wider range. This statement is consistent with the evidence, which mentions that the Mastiff is distinguishable by a limited range of colors. Therefore, I agree with the factuality of the answer in the QA pair. \\

\textbf{Skeptic Agent}: \textcolor{c1}{First, let's analyze the opinions of the previous agent. The previous agent states that the Landseer has a limited range of colors, while the English Mastiff has a wider range. This statement is based on the evidence provided, which mentions that the Mastiff is distinguishable by a limited range of colors. However, the evidence does not explicitly mention the color range of the Landseer. Therefore, it is not entirely accurate to claim that the Landseer has a limited range of colors based on the given evidence. Regarding the factuality of the answer in the QA pair, it is important to note that the evidence does not provide specific information about the color range of the Landseer. Therefore, we cannot definitively conclude whether the Landseer has a limited range of colors or not based on the given evidence. Without further information, it is difficult to determine the factuality of the answer in the QA pair. Considering the above analysis, I would rate the error severity as 2 (Moderate Errors) because the previous agents statement is partially supported by the evidence, but it is not entirely accurate due to the lack of specific information about the Landseers color range.} \\

\textbf{Trust Agent}: \textcolor{c1}{I agree with the previous agents analysis of the opinions. The previous agent correctly points out that the evidence provided does not explicitly mention the color range of the Landseer. Therefore, it is not entirely accurate to claim that the Landseer has a limited range of colors based on the given evidence. However, the evidence does mention that the English Mastiff has a limited range of colors, which supports the previous agents statement about the Mastiff having a wider range of colors compared to the Landseer.} \\

\textbf{Leader Agent}: \textcolor{c1}{After analyzing the opinions of the previous agents, it is clear that both agents agree that the evidence provided does not explicitly mention the color range of the Landseer. Therefore, it is not entirely accurate to claim that the Landseer has a limited range of colors based on the given evidence. However, the evidence does mention that the English Mastiff has a limited range of colors, which supports the previous agents statement about the Mastiff having a wider range of colors compared to the Landseer. Regarding the factuality of the answer in the QA pair, it is important to note that the evidence does not provide specific information about the color range of the Landseer. Therefore, we cannot definitively conclude whether the Landseer has a limited range of colors or not based on the given evidence. Without further information, it is difficult to determine the factuality of the answer in the QA pair.Considering the above analysis, I would rate the error severity as 2 (Moderate Errors) because the previous agents statement is partially supported by the evidence, but it is not entirely accurate due to the lack of specific information about the Landseers color range.} \\

\\ \bottomrule
\end{tabular}
\caption{An example of the debate process by \textbf{GPT-3.5-Turbo}}
\label{tab9}
\end{table*}

\begin{table*}[h!]
\begin{tabular}{p{\textwidth}}
\toprule
\textbf{Question}: Does the Landseer or English Mastiff have a limited range of colors?\\

\textbf{Answer}: The Landseer has a limited range of colours, while the English Mastiff has a wider range.\\

\textbf{Evidence}: It is not to be confused with a white and black Newfoundland, which is also often called a landseer.The English Mastiff is a breed of extremely large dog (often known simply as the Mastiff) perhaps descended from the ancient Alaunt and Pugnaces Britanniae, with a significant input from the Alpine Mastiff in the 19th century. Distinguishable by enormous size, massive head, and a limited range of colours, but always displaying a black mask, the Mastiff is noted for its gentle and loving nature.\\

\textbf{Initial Opinion}: Based on the evidence provided, it is stated that the Landseer has a limited range of colors, while the English Mastiff has a wider range. This statement is consistent with the evidence, which mentions that the Mastiff is distinguishable by a limited range of colors. Therefore, I agree with the factuality of the answer in the QA pair. \\

\textbf{Skeptic Agent}: \textcolor{c2}{The previous agents opinion seems to be slightly misleading. The evidence does not explicitly state that the Landseer has a limited range of colors. It only mentions that the Landseer is often confused with a white and black Newfoundland, which might imply a limited color range, but its not definitive. On the other hand, the evidence does state that the English Mastiff has a limited range of colors, contradicting the QA pairs claim that the English Mastiff has a wider range. Therefore, the factuality of the answer in the QA pair is questionable.} \\

\textbf{Trust Agent}: \textcolor{c2}{While I agree with the previous agents assessment that the evidence does not explicitly state that the Landseer has a limited range of colors, I believe the implication of a limited color range from the confusion with a white and black Newfoundland is a valid interpretation. However, I concur with the previous agents observation that the evidence contradicts the QA pairs claim about the English Mastiffs color range. The evidence clearly states that the English Mastiff has a limited range of colors, which contradicts the QA pairs assertion of a wider range. Therefore, while the QA pairs statement about the Landseer may be inferred from the evidence, the claim about the English Mastiff is factually incorrect based on the provided evidence.} \\

\textbf{Leader Agent}: \textcolor{c2}{The Trust agents interpretation of the Landseers color range being limited due to its confusion with a white and black Newfoundland is a plausible inference, but its not explicitly stated in the evidence. The Skeptic agents doubt about this inference is also valid as the evidence does not directly support it. However, both agents agree that the QA pairs claim about the English Mastiff having a wider color range is contradicted by the evidence, which states that the English Mastiff has a limited range of colors. Therefore, while the claim about the Landseers color range may be subject to interpretation, the claim about the English Mastiff is clearly incorrect based on the provided evidence.} \\

\\ \bottomrule
\end{tabular}
\caption{An example of the debate process by \textbf{GPT-4}}
\label{tab10}
\end{table*}

\begin{table*}[h!]
\small
\textbf{Claim}: Does the Landseer or English Mastiff have a limited range of colors? The Landseer has a limited range of colours, while the English Mastiff has a wider range. \\

\textbf{Evidence}: It is not to be confused with a white and black Newfoundland, which is also often called a landseer. The English Mastiff is a breed of extremely large dog (often known simply as the Mastiff) perhaps descended from the ancient Alaunt and Pugnaces Britanniae, with a significant input from the Alpine Mastiff in the 19th century. Distinguishable by enormous size, massive head, and a limited range of colours, but always displaying a black mask, the Mastiff is noted for its gentle and loving nature. \\

\textbf{Initial state input}: Initial prompt.format(Claim, Evidence)

\textbf{Initial answer}:

\{\\
 "Opinion": "Based on the evidence provided, it is stated that the Landseer has a limited range of colors, while the English Mastiff has a wider range. This statement is consistent with the evidence, which mentions that the Mastiff is distinguishable by a limited range of colors. Therefore, I agree with the factuality of the answer in the QA pair.",\\
 "Factuality": True,\\
 "Error severity": 0\\
\}\\

\textbf{Skeptic agent input}: Skeptic prompt + "Here is the response from the previous agent:" + Initial answer

\textbf{Skeptic agent output}:

\{\\
 "Opinion": "The previous agents opinion seems to be slightly misleading. The evidence does not explicitly state that the Landseer has a limited range of colors. It only mentions that the Landseer is often confused with a white and black Newfoundland, which might imply a limited color range, but its not definitive. On the other hand, the evidence does state that the English Mastiff has a limited range of colors, contradicting the QA pairs claim that the English Mastiff has a wider range. Therefore, the factuality of the answer in the QA pair is questionable.",\\
 "Factuality": False,\\
 "Error severity": 4\\
\}\\

\textbf{Trust agent input}: Trust prompt + "Here is the response from the previous agent:" + Skeptic agent output

\textbf{Trust agent output}:

\{\\
 "Opinion": "While I agree with the previous agents assessment that the evidence does not explicitly state that the Landseer has a limited range of colors, I believe the implication of a limited color range from the confusion with a white and black Newfoundland is a valid interpretation. However, I concur with the previous agents observation that the evidence contradicts the QA pairs claim about the English Mastiffs color range. The evidence clearly states that the English Mastiff has a limited range of colors, which contradicts the QA pairs assertion of a wider range. Therefore, while the QA pairs statement about the Landseer may be inferred from the evidence, the claim about the English Mastiff is factually incorrect based on the provided evidence.",\\
 "Factuality": False,\\
 "Error severity": 4\\
\}\\

\textbf{Leader agent input}: Leader prompt + "Here is the response from the Skeptic agent:" + Skeptic agent output + "Here is the response from the Trust agent:" + Trust agent output

\textbf{Leader agent output}:

\{\\
 "Opinion": "The Trust agents interpretation of the Landseers color range being limited due to its confusion with a white and black Newfoundland is a plausible inference, but its not explicitly stated in the evidence. The Skeptic agents doubt about this inference is also valid as the evidence does not directly support it. However, both agents agree that the QA pairs claim about the English Mastiff having a wider color range is contradicted by the evidence, which states that the English Mastiff has a limited range of colors. Therefore, while the claim about the Landseers color range may be subject to interpretation, the claim about the English Mastiff is clearly incorrect based on the provided evidence.",\\
 "Factuality": False,\\
 "Error severity": 4\\
\}

\caption{Detailed Inference Process}
\label{tab11}
\end{table*}

\end{document}